\documentclass{article}

\usepackage{microtype}
\usepackage{graphicx}
\usepackage{subfigure}
\usepackage{booktabs} 
\usepackage{amsmath}
\usepackage{hyperref}
\usepackage{gensymb}
\usepackage{amsfonts}

\usepackage[accepted]{icml2019}

\icmltitlerunning{PVNet: LRCN for Spatio-Temporal PV Forecast from NWP}

\begin{document}

\twocolumn[
\icmltitle{PVNet: A LRCN Architecture for Spatio-Temporal Photovoltaic Power Forecasting from Numerical Weather Prediction}

\begin{icmlauthorlist}
\icmlauthor{Johan Mathe}{fro}
\icmlauthor{Nina Miolane}{}
\icmlauthor{Nicolas Sebastien}{reu}
\icmlauthor{Jeremie Lequeux}{fro}

\end{icmlauthorlist}

\icmlaffiliation{fro}{Atmo, Inc. San Francisco}
\icmlaffiliation{reu}{Reuniwatt}

\icmlcorrespondingauthor{Johan Mathe}{johan@atmo.ai}
\icmlcorrespondingauthor{Nina Miolane}{nina.miolane@polytechnnique.fr}

\icmlkeywords{PV, Solar, RNN, LSTM, CNN}

\vskip 0.3in
]

\printAffiliationsAndNotice{} 

\begin{abstract}
Photovoltaic (PV) power generation has emerged as one of the lead renewable energy sources. Yet, its production is characterized by high uncertainty, being dependent on weather conditions like solar irradiance and temperature. Predicting PV production, even in the 24-hour forecast, remains a challenge and leads energy providers to left idling - often carbon emitting - plants. In this paper, we introduce a Long-Term Recurrent Convolutional Network using Numerical Weather Predictions (NWP) to predict, in turn, PV production in the 24-hour and 48-hour forecast horizons. This network architecture fully leverages both temporal and spatial weather data, sampled over the whole geographical area of interest. We train our model on an NWP dataset from the National Oceanic and Atmospheric Administration (NOAA) to predict spatially aggregated PV production in Germany. We compare its performance to the persistence model and state-of-the-art methods.

\end{abstract}

\section{Introduction}\label{Intro}


As of early 2019, 184 out of 197 Parties have ratified the Paris Agreement, which was negotiated at the 21st Conference of the Parties (COP21) to reduce the risks and effects of climate change. These signatures show a world-wide awareness of global warming issues and a commitment to planning initiatives that mitigate greenhouse gas emissions. Countries are taking actions to better integrate renewable clean energies into their grids. 

Solar photovoltaic (PV) production is one of the most popular renewable energy sources. According to the International Agency of Energy (IAE) \footnote{https://www.iea.org/topics/renewables/solar/}, cumulative solar PV capacity reached almost 398 GW in 2017, representing around 2\% of global power output. Furthermore, solar PV is expected to lead renewable electricity capacity growth, increasing to over 1000 GW by 2023. 

Integrating solar PV production into the energy grids requires an accurate and reliable forecast of the PV power, in order to guarantee the optimal management of energy demand and supply. Accurate PV power forecasting remains a challenge as the PV output depends on fluctuating weather conditions, like solar irradiance or other meteorological factors \cite{Li2016a}. 

\subsection{Related Works}

Many studies have tackled the challenge of PV power forecast. They rely on statistical time-series methods, physical methods, or ensemble methods which combine different models to enhance accuracy \cite{Sobri2018}. Widely used time-series models can be separated into two groups: linear and non-linear models. Linear time-series models used in PV forecast include the combination of auto-regressive models (AR) with moving average models (MA) into auto-regressive moving average models (ARMA) \cite{Li2014AnSystem, Kardakos2013}. While linear models are usually more transparent towards their feature selection, non-linear models often show better prediction accuracy when enough data are available. Among non-linear time-series models, Artificial Neural Networks (ANN) have become increasingly popular for PV forecast \cite{Ding2011AnSystem, Kardakos2013, Dolara2015} and among them Recurrent Neural Networks \cite{Malvoni2013} and Long-Short Term Memory Networks \cite{Abdel-Nasser2017, Gensler2017DeepNetworks}. We note that ANN models can be interpreted as a non-linear version of AR, also called NAR, while Recurrent Neural Networks can be seen as a type of non-linear ARMA, also called NARMA. 

The most successful methods among the ones mentioned above are models adding Numerical Weather Prediction (NWP) like predictions of solar irradiance or temperature into the time-series approach. They may also include analytical physical model outputs, like atmospheric models and several types of Clear Sky Models \cite{Inman2013SolarIntegration}. NWP can be obtained from the European Centre for Medium-Range Weather Forecasts (ECMWF) \footnote{https://www.ecmwf.int/en/forecasts/datasets/set-i} and the National Oceanic and Atmospheric Administration (NOAA) \footnote{https://www.emc.ncep.noaa.gov/GFS/doc.php}. Clear Sky Models and additional physical models are provided by open-source libraries \cite{Holmgren2018PvlibSystems}. Time-series models including these additional inputs are referred to as ARMA model with exogenous inputs (ARMAX) for the linear versions \cite{Kardakos2013} or NAR with exogenous inputs (NARX) \cite{DiPiazza2016} as well as Artificial Neural Networks with multiple inputs \cite{Malvoni2013, Dolara2015, Hossain2018, Oneto2018}. 

Adding NWP and physical model inputs to time-series models show better forecast prediction, yet limited computational resources as well as dataset sizes often prohibit studies to integrate an exhaustive list of these variables, especially in non-linear models. Instead, the impact of the variables is assessed by independent experiments \cite{Kardakos2013} of by fitting regression models like multi-regression analyses \cite{Malvoni2013} or multivariate adaptive regression splines \cite{Li2016a}. Eventually, a subset of the variables may be selected for prediction. These variable selection analyses are conducted at specific locations, at four PV plants in Greece \cite{Kardakos2013}, one PV plant in Southern Italy \cite{Malvoni2013} and in Macau \cite{Li2016b}, while the variables impact on the prediction is likely to be location-dependent.

In this paper, we present a method that takes into account the spatial distribution of the input variables, to simultaneously increase the model's inference capacity - an understanding of which inputs variables are essential at which geographic locations - as well as the model's prediction accuracy. In the above studies, the spatial distribution of the NWP and physical models variables are not taken into account: models focus for example on solar irradiance or temperature only at the specific location of the target plant. The time-series and neural network architectures often model the temporal structure of the variables, not their spatial structure and initiatives to incorporate spatial information are often restricted to sparse data at neighboring plants for example \cite{Gutierrez-Corea2016, Agoua2018}. Yet, weather variables and physical model outputs on the zone surrounding a target plant are additional data, especially valuable while fitting non-linear flexible models, that can be integrated into the input data for higher accuracy. To the best of our knowledge, no model fully integrates both temporal and dense spatial numerical weather predictions and physical models outputs in PV forecasting. 

Meanwhile, flexible non-linear predictions models taking into account the spatiotemporal structure of the data, like Long-Term Recurrent Convolutional Network (LRCN) \cite{Donahue2015}, 2D LSTM or Convolutional LSTM architectures (ConvLSTM) \cite{Shi2015}, have been successfully applied to a variety of problems. LRCN models have been used in activity recognition, image captioning and visual question answering \cite{Donahue2015}. A 2D LSTM model has been applied to traffic forecasting \cite{Zhao2017LSTMForecast} while a ConvLSTM has shown promising results on a precipitation forecast that predicts rainfall intensity in a local region on a  short time horizon \cite{Shi2015} (also known as nowcasting), a weather-related prediction problem that shares similarities with PV forecast. 

\subsection{Contributions and Outline}

This paper presents PVNet, a PV forecasting model that introduces a LRCN architecture to integrate past PV power 1D time-series with dense spatiotemporal NWP and physical models' inputs. Our model extends the state-of-the-art 1D time-series models by learning spatial features with CNN modules, whose channels encode the NWP and physical models variables. We focus our analysis on the day-ahead PV forecast, whose accuracy is known to be highly dependent on NWP integration, and which is particularly interesting as energy market bids are placed one day in advance. Our experimental results show that our architecture manages to leverage these additional spatial data for prediction accuracy. We train a highly non-linear prediction model that reaches a high accuracy on our validation dataset. Furthermore, our model provides a location-dependent assessment of the impact of the different input variables, through an occlusion sensitivity analysis that shows a strong inference capability.

In Section~\ref{sec:model}, we describe the data representation used for the PV power output and the NWP and physical models variables. We then present our spatiotemporal LRCN model PVNet, with its CNN and LSTM modules. Section~\ref{sec:exp} describes our experimental studies and accuracy results, as well as the results of the occlusion sensitivity analysis that validates our spatiotemporal approach.

\section{Data Representation\label{sec:model}}

\subsection{Photovoltaic Panel}

\begin{figure}[ht]
\vskip 0.2in
\begin{center}
\includegraphics[width=\columnwidth]{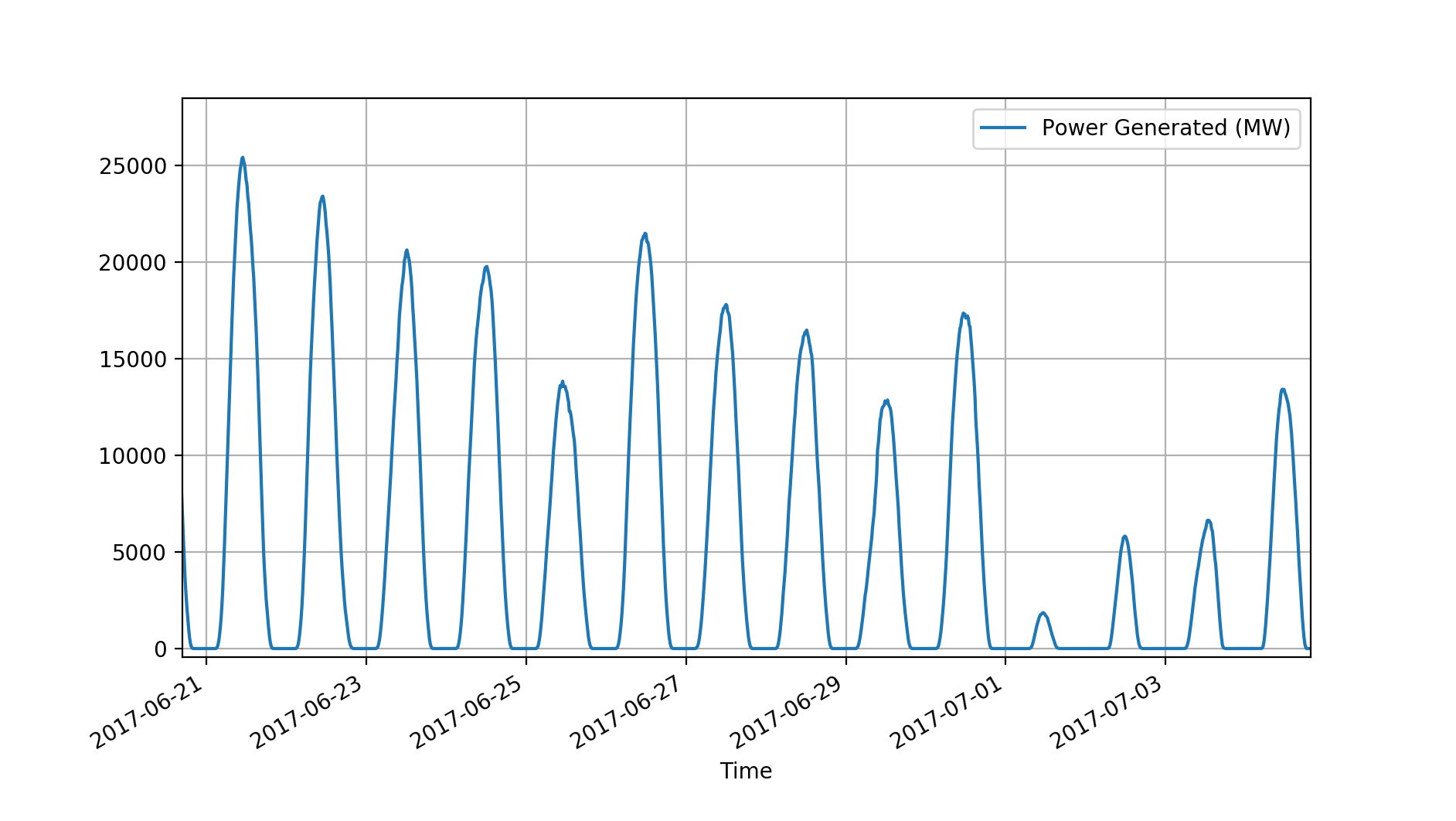}
\caption{Aggregated power over a few days\label{fig:pv}}
\end{center}
\vskip -0.2in
\end{figure}

A photovoltaic solar panel, also called photovoltaic module, is a device that absorbs sunlight as a source of energy to generate electricity. We are interested in the power output produced by the module, which is a time series that depends on weather conditions, see Figure~\ref{fig:pv}.

\begin{figure}[ht]
\vskip 0.2in
\begin{center}
\includegraphics[width=3cm]{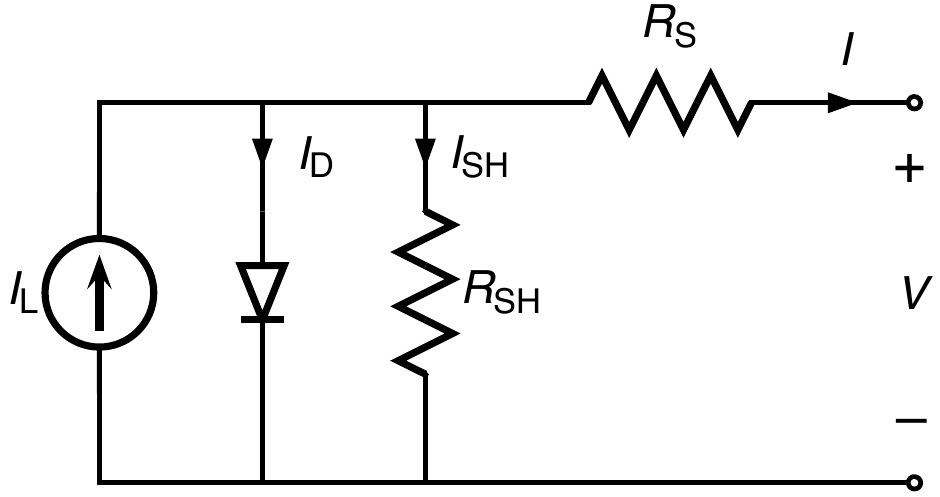}
\caption{Equivalent Circuit of a Solar Cell.} \label{fig:circuit}
\end{center}
\vskip -0.2in
\end{figure}

There are different ways to model the solar panel production from external factors, one of which is the equivalent electrical circuit of a solar cell is composed of a diode, a resistor in series and in parallel as shown in see Figure~\ref{fig:circuit} \footnote{\href{https://commons.wikimedia.org/wiki/File:Solar_cell_equivalent_circuit.svg}{https://commons.wikimedia.org}}). 

The diode current is given by the Shockley diode equation:

\begin{equation}
    I_{D} = I_0 (e^{\frac{V_j}{nV_{T}}} - 1)
\end{equation}

With $V_T$:
\begin{equation}
V_T = \frac{kT}{q}
\end{equation}
With $k$ the Boltzmann constant, $T$ the absolute temperature of the p-n junction, and $q$ the elementary charge.

Thus the output power produced by the module obviously depends on solar irradiance, but also on its temperature \cite{Fesharaki2011TheEfficiency}. 

\begin{equation}
    P_{pv} = V_{pv} \times I_{pv} 
\end{equation}

\begin{equation}
    I_{pv} = I_{l} - I_{0}(e^{q(V_{pv}+I_{pv}R_{s})/AKT}-1) -\frac{V_{pv}+I_{pv}R_{s}}{R_{sh}}
\end{equation}

Furthermore, recent experimental research suggests that a module's temperature is related to numerical weather data such as the ambient temperature, the local irradiance and wind speed. \cite{Muzathik2014PhotovoltaicCorrelation}:

\begin{equation}
    T_{\text{m}} (\degree\text{C}) = 0.94  T + 0.02 \text{I} - 1.5 \text{S} + 0.35
\end{equation}
where $T_{\text{m}}$ is the module's temperature, $T$ is the \textit{ambient} temperature, $I$ is the solar irradiance and $S$ the wind speed. However, this model essentially comes from an heuristic and there are likely more weather data - and their non-linear relationships - that could be taken into account to fully characterize the module's temperature and hence, its power output.

We just showed that the behavior of a photovoltaic panel is both non linear because of its semi-conductor nature and also dependant on the temperature of the panel. Temperature which is itself dependant on external factors like external temperature and windspeed which are available and forecast through numerical  services. 

\subsection{Numerical Weather Predictions}

We incorporate weather prediction data as well as their non-linear relationships in our model. Rather than just incorporating the irradiance and temperature at the specific location of the power plant(s), our model aims to leverage the full spatial scalar fields of the meteorological parameters over the whole area of interest. These forecasts can be obtained from global forecast systems, like the ECMWF HRES model (temporal resolution of 1 hour and spatial resolution of 0.1 degree) \footnote{\href{https://www.ecmwf.int/en/forecasts/datasets/set-i}{https://www.ecmwf.int/en/forecasts/datasets/set-i}} or the NOAA GFS \footnote{\href{https://www.emc.ncep.noaa.gov/GFS/doc.php}{https://www.emc.ncep.noaa.gov/GFS/doc.php}} (temporal resolution of 3 hours and spatial resolution of 0.5 degrees). Each of these models provides a set of weather-related variables.

\begin{figure}[ht]
\vskip 0.2in
\begin{center}
\includegraphics[width=4cm]{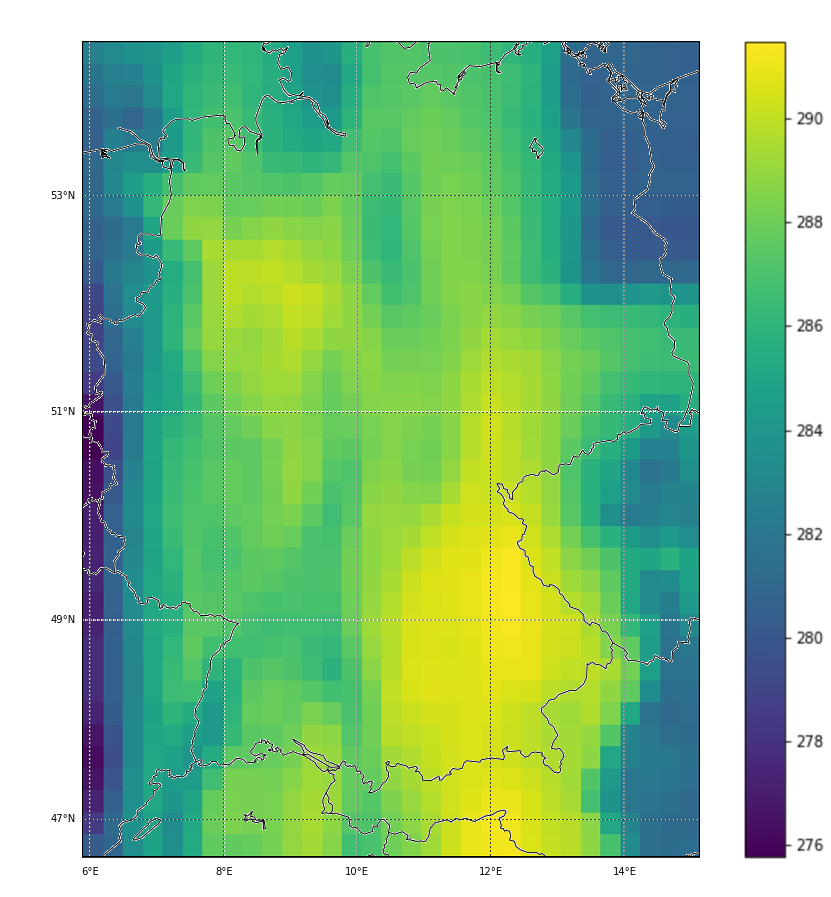}
\vspace{2em}
\includegraphics[width=4cm]{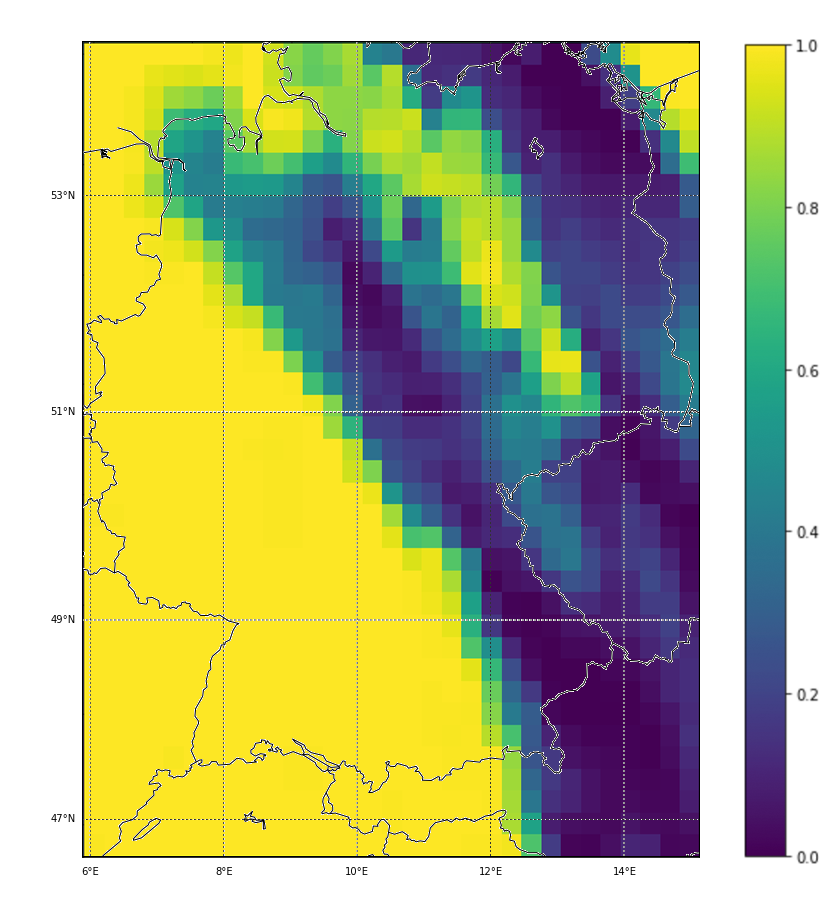} \\
\caption{Temperature (in Kelvin) and Cloud Cover (unitless) maps over Germany on May 1st, 2016}
\label{fig:nwp}
\end{center}
\vskip -0.2in
\end{figure}

Figure~\ref{fig:nwp} shows examples of variables of interest, in this case over our country of interest Germany.

\begin{figure}[ht]
\vskip 0.2in
\begin{center}
\centerline{\includegraphics[width=\columnwidth]{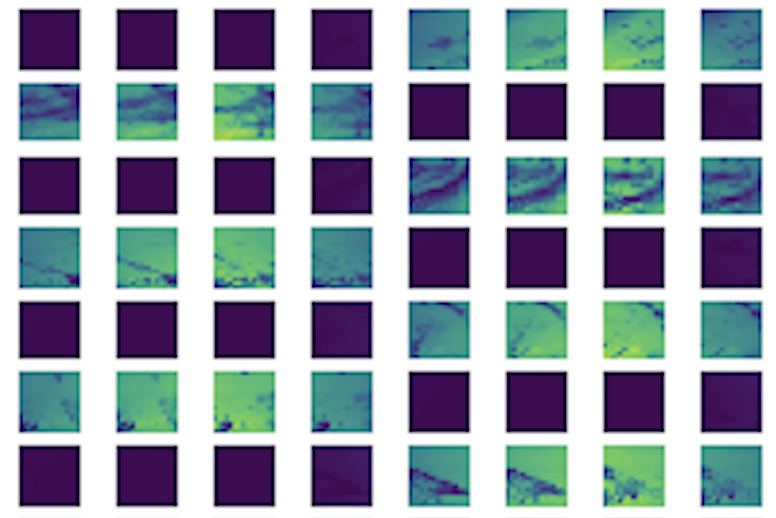}}
\caption{NOAA GFS Irradiance over Germany for 7 days - one patch every 3 hours - time goes from left to right. Lighter color values are higher irradiance. High level view show alternance of days and nights.\label{fig:irradiance}}
\end{center}
\vskip -0.2in
\end{figure}

Figure ~\ref{fig:irradiance} shows a collection of weather patches that typically represents input data. Here one patch every three hours.

In order to predict PV power output in the day-ahead forecast, we use the numerical weather predictions of the above variables for this time horizon. 
\begin{equation}\label{eq:feature}
    X_{NWP, t}(x, y) = 
        \begin{pmatrix}
            NWP_1(x, y)\\
            \vdots \\
            NWP_K(x, y)
        \end{pmatrix}
\end{equation}
where $t$ is the time, measured in hours from the current time, $(x, y)$ is the longitude-latitude coordinates of a point in the region of interest, and $K$ is the number meteorological factors of interests.

We chose this set of weather features after investigating the intersection of various available data in the NWP forecast datasets and the current state of the art for the impact of the variables affecting the irradiance (here cloud cover and irradiance) and the performance of the solar panel (here outside temperature).

\subsection{Incorporation of Standard Models Predictions}

\begin{figure}[ht]
\vskip 0.2in
\begin{center}
\centerline{\includegraphics[width=\columnwidth]{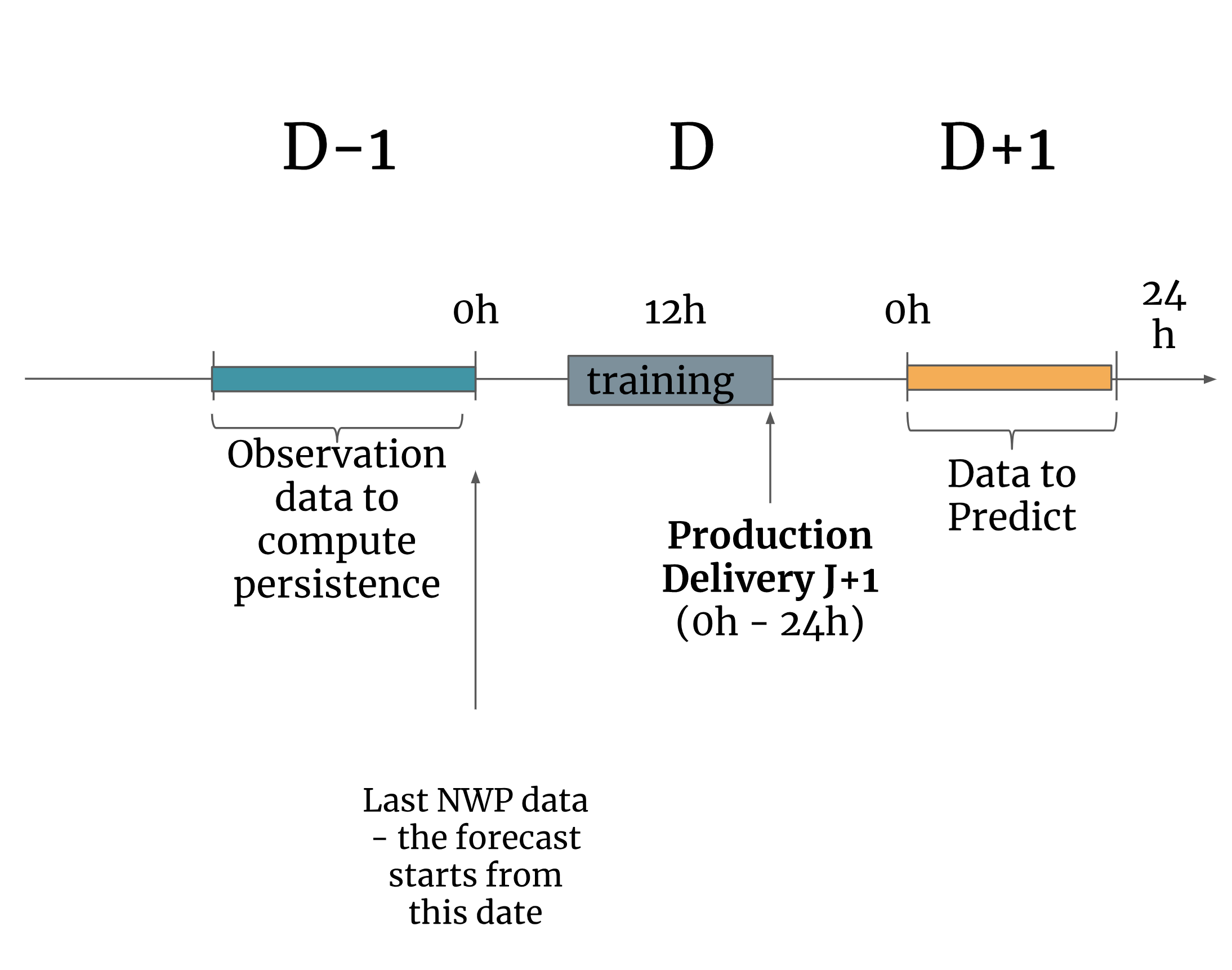}}
\caption{Timeline of day ahead prediction\label{fig:timing}}
\end{center}
\vskip -0.2in
\end{figure}

We also incorporate predictions of standard forecast models into our feature vector. We consider first the persistence model, a simple - yet popular - forecast model. The persistence model predicts that future PV output is likely to be similar as the current value at the same time of the day \cite{Cornaro2015}. We denote with the subscript $PSS$, the PV output predicted by a persistence model:
\begin{equation}
    X_{\text{PSS}}(x, y, t) = P_{pv}(t - T_{p})
\end{equation}
where time $T_{p}$ is measured in hours, and $(x, y)$ the longitude-latitude coordinates. Here we take two specific values of persistence. One will be used for comparing the performance of the algorithm at $T_{24}$ and the one we use as a feature of our algorithm is at $T_{48}$. Indeed when used in the context of energy forecast, there is a 24 hours delay, as seen in figure~\ref{fig:timing}.

A clear sky model estimates irradiance under the assumption that there are no clouds in the sky \cite{RenoSANDIAAnalysis}. This is a function of the altitude, location, and various atmospheric conditions. The irradiance predicted by the clear sky model is:
\begin{equation}
    X_{\text{CSM}, t}(x, y) = \text{CSM}(t, x, y, \alpha(t, x, y))
\end{equation}
where $t$ is the time, measured in hours from the current time, $(x, y)$ is the longitude-latitude coordinates of a point in the region of interest and $\alpha$ is a set of external variables used for the Clear Sky Model. 

Specifically, we form a feature at time $t$, $X_t$ by aggregating the weather forecast variables with the predictions of the standard models:

\begin{equation}\label{eq:feature2}
    X_t(x, y) = 
        \begin{pmatrix}
            X_{\text{NWP}, t}(x, y)\\
            X_{\text{PSS}, t}\\
            X_{\text{CSM}, t}(x, y)
        \end{pmatrix} \in \mathbb{R}^{K+2}
\end{equation}
where $t$ is the time, measured in hours from the current time, $(x, y)$ is the longitude-latitude coordinates of a point in the region of interest. thus, our feature vector incorporates both spatial weather forecasts as well as some predictions from existing models.

\section{PVNet Model}

The goal of PV power output forecasting is to use numerical weather predictions, to forecast the future PV output in a region, in our case: at the day ahead forecast at the scale of a country. We consider $t=0$ the time at which we make the prediction, and thus $t_{\text{forecast}}=24h$ is the forecast time. Precisely, we consider a sliding window of size $T$, and:
\begin{equation}
    \left(X_{t_1}, ..., X_{t_T} \right)
\end{equation}  
where $X_t$ is the feature maps at time $t$, as defined in Equation~\ref{eq:feature2} and $t_T = t_{\text{forecast}}$. Our goal is to predict:

\begin{equation*}
    \widehat{P_{pv}}(t_T),
\end{equation*}

an estimate of the PV production at the forecast horizon. Our training set is thus a set of $\left(\left(X_{t_0 + t_1}, ..., X_{t_0 + t_T} \right), P_{pv}(t_0 + t_T)\right)_{t_0}$ for different values of $t_0$.

\begin{figure}[ht]
\vskip 0.2in
\begin{center}
\centerline{\includegraphics[width=\columnwidth]{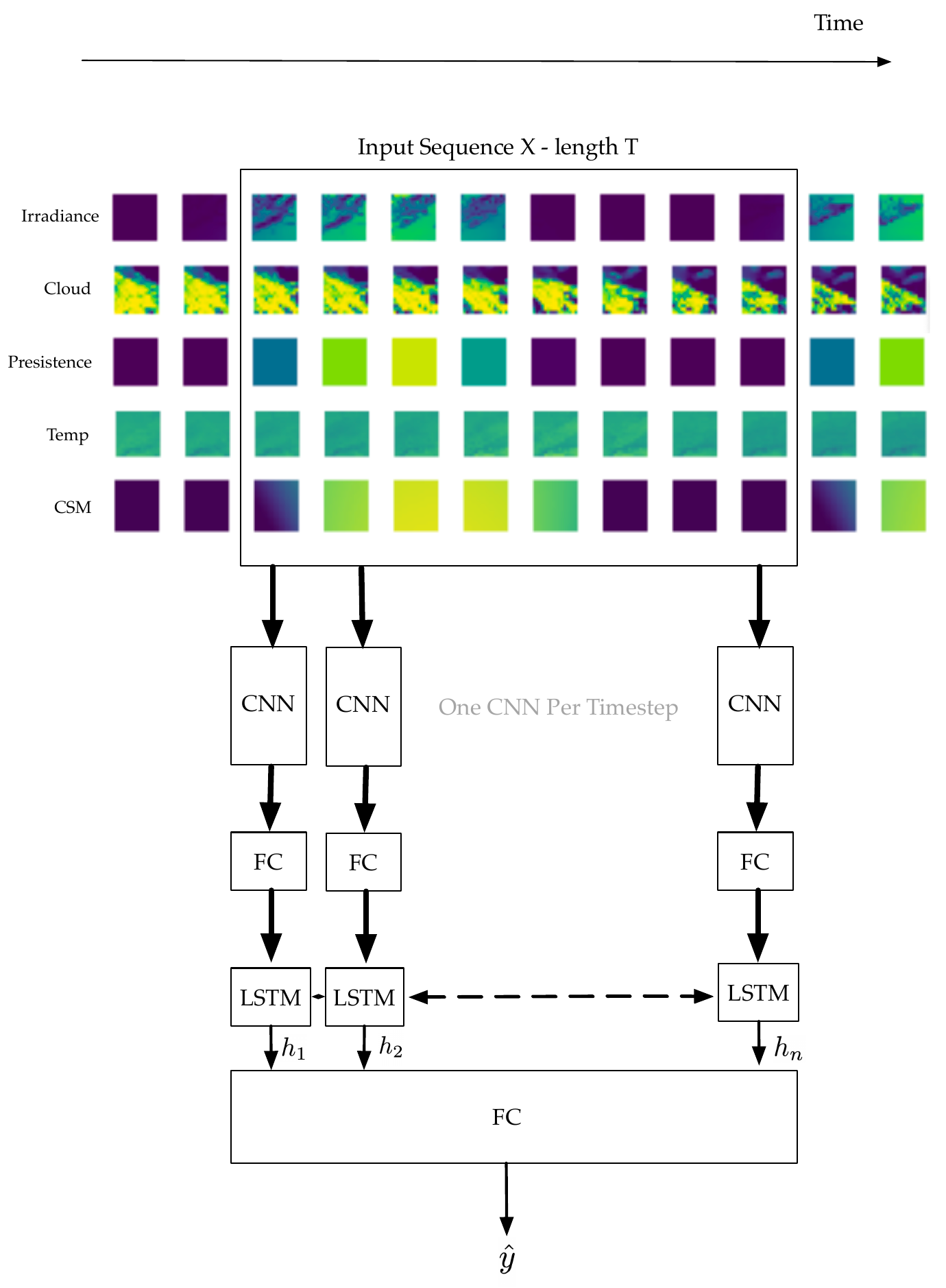}}
\caption{PVNet Architecture Details\label{fig:network}}
\label{icml-historical}
\end{center}
\vskip -0.2in
\end{figure}

Our model presents a long term recurrent convolutional network which combines a convolutional network (CNN) and a bi-directional long short term memory (LSTM) \cite{Schuster1997BidirectionalNetworks} network, as shown in Figure~\ref{fig:network}. In practice, we use a time window of $T=8$ values, spread at 3 hours time interval, so as to represent a window that is a full day of data history.

\subsection{The CNN Module}

The model applies a convolutional neural network to each $T$ images $X_t$ of shape (width, height) and $K+2$ number of channels. 
This results into a feature vector of dimension (T, d):
\begin{equation}
\left(x_{t_1}, ..., x_{t_T} \right) = \left(g(X_{t_1}), ..., g(X_{t_T}) \right)
\end{equation}
where each $x_t$ is now a feature vector of dimension $d$. We use one single CNN architecture, and a single set of weights, to transform the feature $X_{t}$, independently of the time $t$. Therefore, this feature transformation step can be described by a single function $g$.

We use the CNN module to encode the spatial data into a new feature vector of lower dimension as running the LSTM directly on the whole images would be too computationally expensive. Rather than hand-crafting spatially aggregated features, we let the CNN learn spatial features that are the most relevant for the following LSTM-based PV power predictions.

\subsection{The LSTM Module}

Each of the feature tensor is then fed into an LSTM \cite{Hochreiter1997} network, to incorporate the time-series information. A LSTM network is a special case of Recurrent Neural Network (RNN), which introduces a memory cell $c_t$, that serves as an accumulator of state information. The cell $c_t$ can be accessed, written and cleared by self-parameterized controlling gates. If the input gate $i_t$ is activated, the information of each new input, that is our feature vector $x_t$, is accumulated to the cell. If the forget gate $f_t$ is activated, the previous cell status $c_{t-1}$ can be forgotten. The output gate $o_t$ controls if the cell output $c_t$ will be propagated to the final state $h_t$. The model can be summarized with the following equations:

\begin{equation}
\begin{split}
f_{t} &= \sigma _{g}(W_{f}x_{t}+U_{f}h_{t-1}+b_{f})\\
i_{t} &= \sigma _{g}(W_{i}x_{t}+U_{i}h_{t-1}+b_{i}) \\
o_{t} &= \sigma _{g}(W_{o}x_{t}+U_{o}h_{t-1}+b_{o}) \\
c_{t} &= f_{t} \circ c_{t-1}+i_{t}\circ \sigma _{c}(W_{c}x_{t}+U_{c}h_{t-1}+b_{c})\\
h_{t} &= o_{t} \circ \sigma _{h}(c_{t})
\end{split}
\end{equation}

One of the limitations of the LSTM is that any hidden layer associated with a timestep $t$ only has access to past data. Bidirectional RNNs \cite{Schuster1997BidirectionalNetworks} have shown to be more efficient in natural language processing tasks \cite{Arisoy2015BidirectionalRecognition}. The following equations show how the RNN takes into account past and future data:

\begin{equation}
\begin{split}
\overrightarrow{h_{t}} &= f(\overrightarrow{W}x_t + \overrightarrow{V} \overrightarrow{h_{t-1}} + \overrightarrow{b}) \\
\overleftarrow{h_{t}} &= f(\overleftarrow{W}x_t + \overleftarrow{V} \overleftarrow{h_{t+1}} + \overleftarrow{b}) \\
\hat{y}_t &= g(Uh_t+c) = g(U[\overrightarrow{h_t};\overleftarrow{h_t}] + c) 
\end{split}
\end{equation}

The intent of the BiLSTM module is to design features, at each time $t$ that take into account the temporal structure of the input signal.

\subsection{Full Model}

The convolutional network extracts spatial features (in our case spatial weather related features) and then forwards them to the LSTM. We then have a fully connected layer taking all the LSTM hidden outputs and producing a single scalar value that is the estimated value of aggregated PV output prediction for the whole country.

We highlight that this model differs from the ConvLSTM model from \cite{Shi2015}, in the sense that the ConvLSTM model integrates convolutional structures directly in the LSTM cells.

Both CNN and LSTM modules are trained simultaneously, to minimize the mean squared error loss function:
\begin{equation}
    MSE = \frac{1}{N} \sum_{t=t_0}^{t_{N-1}} \left( \widehat{P_{pv}}(t + t_T) - P_{pv}(t + t_T)\right)^2,
\end{equation}
summed over windows' initial time values $t_0$.

\subsection{Implementation Details}

We implement PVNet using the Tensorflow \cite{AbadiTensorFlow:Systems} framework. Our convolutional networks are built alternating 2D convolution layers, PReLu activations \cite{HeDelvingClassification}, dropout and max pooling.

There is also a 20\% dropout after each convolutional layer. We added a fully connected layer between the CNN layers and the LSTM, with a 30\% dropout in between.

In this implementation, we use the hard sigmoid for the recurrent activation and the hyperbolic tangent for the activation itself.

\begin{table}[!ht]
\caption{CNN Layers Details\label{tab:results2}}
\vskip 0.15in
\begin{center}
\begin{small}
\begin{sc}
\begin{tabular}{lcccr}
\toprule
Layer  & Parameters & Activation\\
\midrule
Conv2D  & 3x3x64      & PReLu \\
Conv2D    & 3x3x64      & PReLu \\
MaxPool    & 2x2   & N/A \\
Conv2D    & 3x3x128      & PReLu \\
Conv2D    & 3x3x128     & PReLu \\
MaxPool       & N/A \\
Conv2D    & 3x3x256     & PReLu \\
Conv2D     & 3x3x256     & PReLu \\
MaxPool       & 2x2 \\
Flatten     & N/A  & N/A \\
FC       & 512      & N/A \\
\bottomrule
\end{tabular} 

\end{sc}
\end{small}
\end{center}
\vskip -0.1in
\end{table}

\section{Dataset}

We consider an aggregated production of PV power across Germany. We gather past data from 2014 to 2018, sampled at a time resolution of 15 minutes.

We use numerical weather prediction data from the National Oceanic and Atmospheric Administration, more specifically the Global Forecast System (GFS). This data is freely available, has a spatial resolution of 0.5 degrees of latitude and longitude and a time resolution of 3 hours. The time resolution of the PV power is downsampled at 3h, to match the time resolution of the NWP. We consider $K=3$ meteorological variables: Downward short wave radiation flux (DSWRF), Cloud Cover (EACC) and Temperature (TMP). We incorporate data from the persistence model and the Clear Sky Model.

\subsection{Time Alignment}

Time alignment is critical for efficient training. When first looking at the GFS data one has to be careful about time shifts. The persistence data included in the model is not the one used to evaluate the performance of the model. Indeed in a production setting one only has access to the data from the day before. This means that we can only use persistence data from 48 hours before the point to estimate, whereas the actual persistence metrics is done 24 hours before.

\subsection{Training and Validation Split}

We split the training and validation dataset as follows: first 3 years for training and the last year for validation. This approach guarantees that our validation score is not affected by seasonal patterns. Indeed, the duration of the day is very different between the summer and the winter solstice, performance measures could be affected without this specific care.

\section{Experiments and Results\label{sec:exp}}

\begin{figure}[ht]
\vskip 0.2in
\begin{center}
\centerline{\includegraphics[width=\columnwidth]{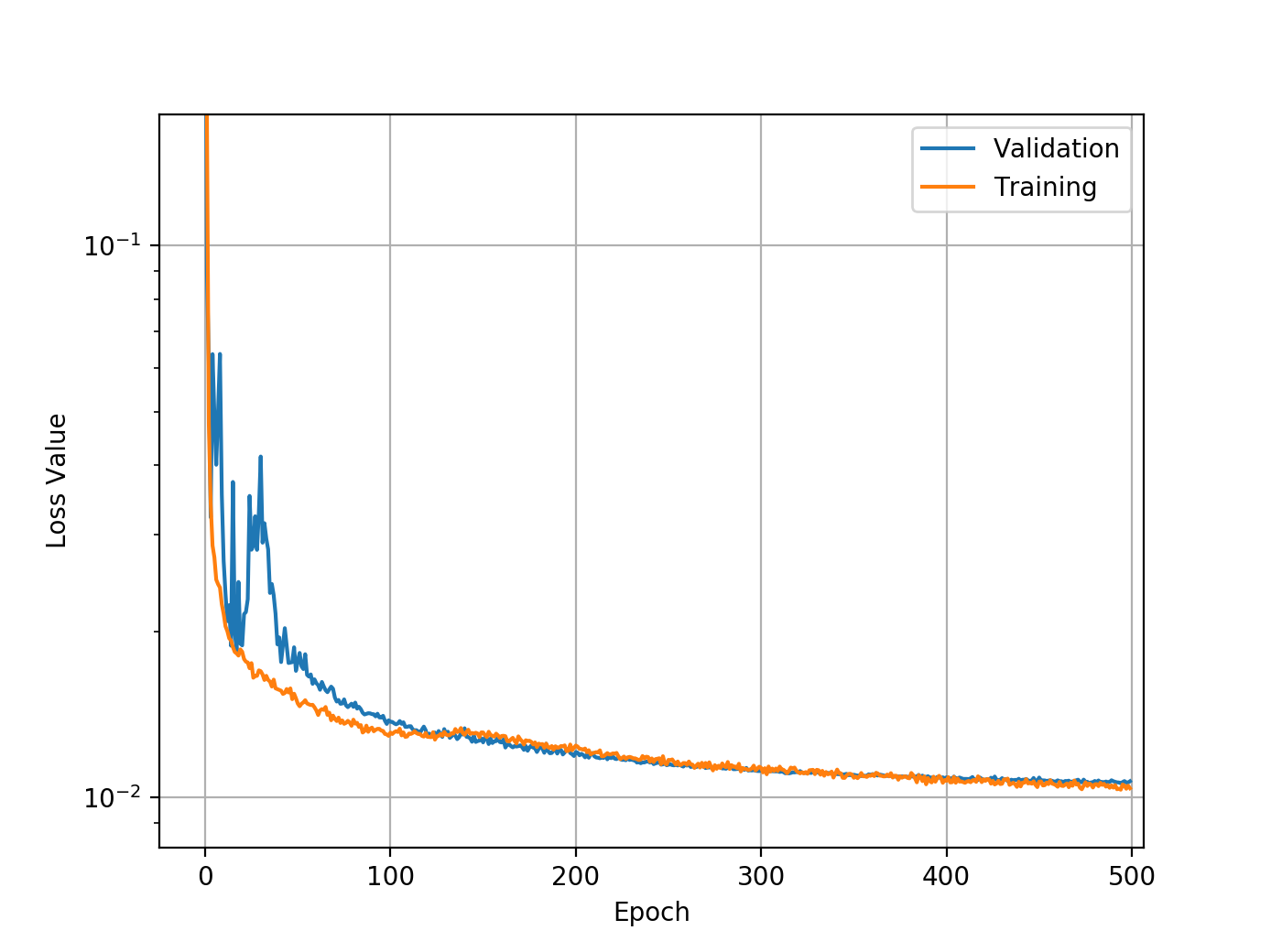}}
\caption{PVNet Training and Validation Losses for n=500 epochs.\label{fig:losses}}
\end{center}
\vskip -0.2in
\end{figure}

We run our experiments on a computer with a single NVIDIA Titan X GPU. The training time of the network is 5 hours. We train our model using the Adam optimizer \cite{Kingma2014}, with a learning rate of 0.0015 and a batch size of 32. We run our training for a total of 500 epochs, as shown in Figure~\ref{fig:losses}.

\subsection{Metrics and Results}

We use two metrics to evaluate the quality of PVNet: the root mean squared error (RMSE) and mean absolute error (MAE) of the prediction.  We compute these metrics only when the measured and predicted power is higher than 0 ($P_{PV}(t) > 0$). We do not take nights into accounts because these values tend to increase the performance since the estimator of the PV night is $\hat{P_{PV}}(t) = 0$ The normalized RMSE and MAE (nRMSE and nMAE) are normalized with the country-wide PV capacity, which is 41.2 GW in our case.

Tables~\ref{tab:results2} and \ref{tab:results} present our results for PVNet compared to the persistence model and the current state of the art for day ahead country wide PV prediction \cite{Lorenz2011LocalDetection}. \cite{Lorenz2011LocalDetection} used data from 2009-2010 and a different formulation of the nRMSE. The authors of \cite{Lorenz2011LocalDetection} indeed normalized the production on a per plant basis while we normalize it at the scale of the whole country. We re-implement their method in order to compute the metrics MAE, nMAE, RMSE and nRMSE for the same times as the ones used in our validation dataset.

\begin{table}[h!]
\caption{PVNet Experimental Performance\label{tab:results}}
\vskip 0.15in
\begin{center}
\begin{small}
\begin{sc}
\begin{tabular}{lcccr}
\toprule
Method  & nRMSE (\%)   & nMAE (\%)\\
\midrule
Persistence        & 22.04 & 15.28 \\

Lorenz et al     & 6.11   & 4.37  \\
\textbf{Pvnet}              & \textbf{4.73}   & \textbf{3.63}  \\
\bottomrule
\end{tabular} 
\\
\vspace{2em}

\begin{tabular}{lcccccr}
\toprule
Method & RMSE (MW)  & MAE (MW)  \\
\midrule
Persistence      & 8816   & 6297  \\

Lorenz et al   & 2518    & 1798   \\
\textbf{Pvnet}            & \textbf{1949}   & \textbf{1499}   \\
\bottomrule
\end{tabular}
\end{sc}
\end{small}
\end{center}
\vskip -0.1in
\end{table}

We compute the persistence model RMSE by taking the value of the country-aggregated prediction 24 hours before the current value. We first observe an RMSE improvement of 17.31 percents compared to the persistence model, which is expected since the persistence model is relatively naive. We do also use the persistence data in our model as one of the inputs - even though the timing is quite different. We use 24 hours persistence for verification and 48 hours persistence as the PVNet input layer. 

We also observe a 1.38\% decrease in accuracy error in nRMSE and a 0.74\% decrease in accuracy error in nMAE compared to state of the art in the country-wide day ahead aggregation \cite{Lorenz2011LocalDetection}. Non-linear spatiotemporal features are being captured in much more details in PVNet, allowing better overall performance.

\subsection{Occlusion Sensitivity Analysis\label{sec:diagnosis}}

One of the main interests of this model is to build a spatial representation of the impact of various variables (weather, clear sky model and persistence) on the aggregated PV production. One way to verify if this is the case is to perform a sensitivity analysis to evaluate the spatial impact of each variable.

This type of analysis has been shown to work well in the case of deep neural networks \cite{Zeiler2013VisualizingNetworks}.

\begin{figure}[ht]
\vskip 0.2in
\begin{center}
\centerline{\includegraphics[width=\columnwidth]{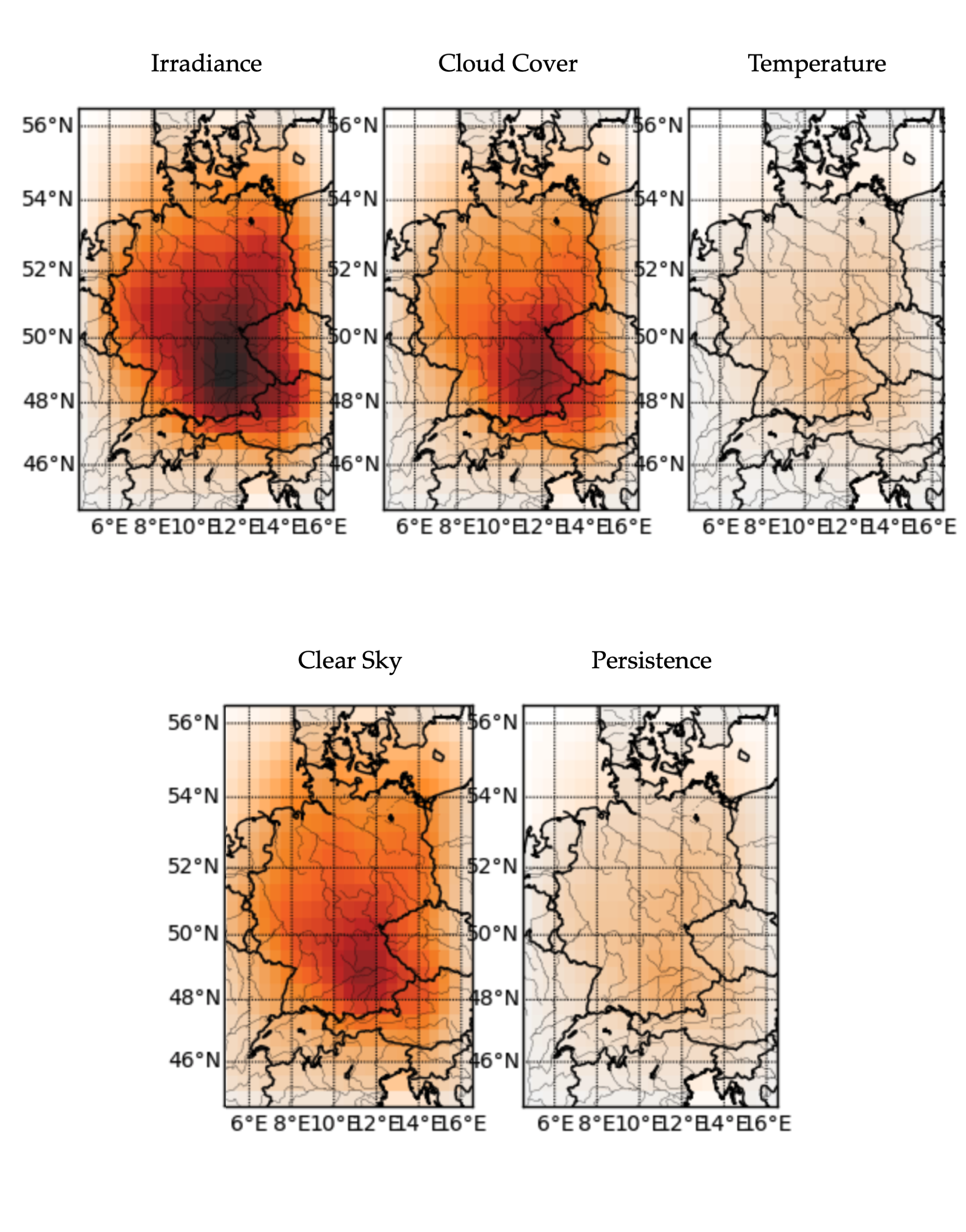}}
\caption{Occlusion Sensitivity Analysis for a 2x2 Patch. Black is highest sensitivity, white/light orange is lowest sensitivity.\label{fig:sensitivity}}
\label{deconv}
\end{center}
\vskip -0.2in
\end{figure}

We perform an occlusion sensitivity analysis by setting the values of a patch to a fixed value and recording the changes in value of the predicted power prediction. We perform this analysis for each channel and aggregated the values in a heatmap form. We then overlay this heatmap with the map of Germany and we obtain the results as shown in \ref{fig:sensitivity}. The results show that our network sees the biggest impact with irradiance, followed by cloud cover, clear sky, persistence and finally temperature.

\begin{figure}[ht]
\vskip 0.2in
\begin{center}
\centerline{\includegraphics[width=5cm]{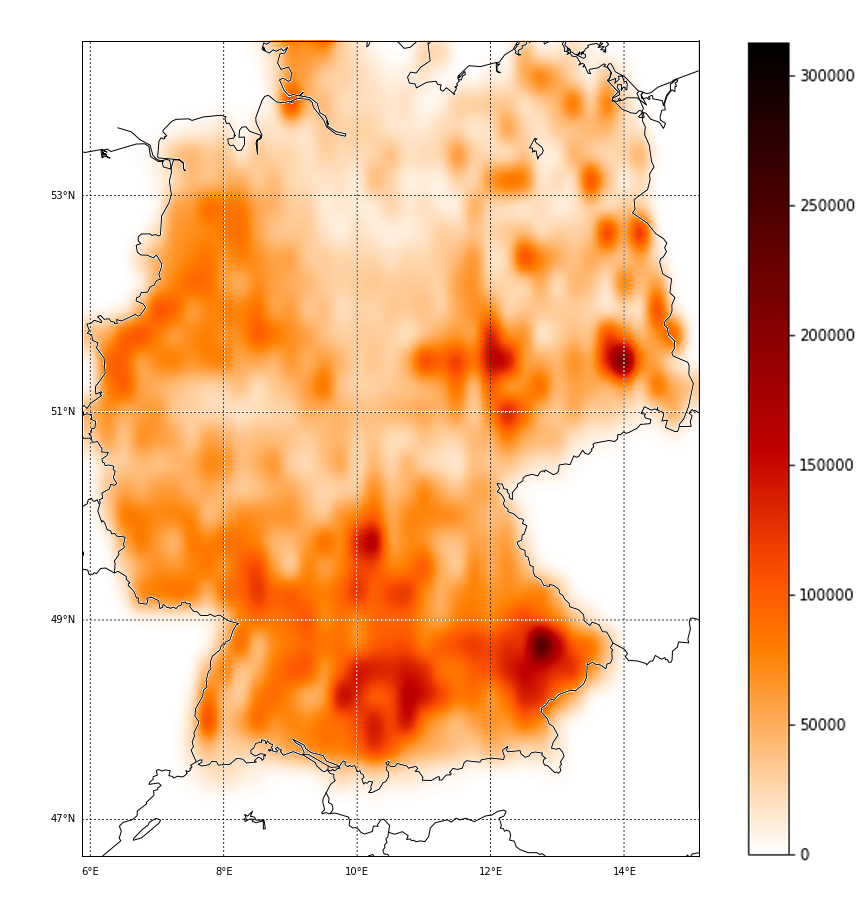}}
\caption{Distribution of PV power generated across Germany. The unit is power in Watts produced per area of $0.25\times0.25$ degrees of latitude and longitude.\label{fig:plants}}
\label{deconv}
\end{center}
\vskip -0.2in
\end{figure}

In order to compare the results of the sensitivity analysis we build a heatmap representing the distribution of PV power generated across Germany on May 1st, 2016. We aggregate the nominal capacity of the plants on a spatial grid of $0.25 \times  0.25$ degrees by summing all maximum plant power on this area. We overlay this heatmap with Germany country boundaries for more clarity (see figure \ref{fig:plants}). This shows that the south/east of Germany produces more power per squared meter, which is consistent with the results of our occlusion map. Indeed most of the metrics in the sensitivity maps (figure~\ref{fig:sensitivity}) show higher incidence in the south west. 

This sensitivity analysis also shows which input features have the biggest impact on the outputs. Indeed we that in order or decreasing importance, the values are irradiance, followed by cloud cover, the clear sky channel and finally Persistence and Temperature. This analysis also gives us a good insight about PVNet - looking at these two figures we can conclude that applying PV net and a Sensitivity analysis can be used to learn an estimate of the surface density of PV power production for a given area.

\section{Conclusion and Future Work}

Global warming and the current energy crisis are both calling for the integration of renewable energies, including solar PV energy, into countries power grids. Such integration is still limited by our ability in making reliable PV power forecast. Accurate PV forecast remains challenging because of its correlation with highly fluctuating weather variables.

This paper introduced PVNet, a country-wide PV forecast model that fully integrates 1D time-series of past PV power with dense spatiotemporal exogenous inputs. Specifically, PVNet leverages an LRCN architecture that tackles spatiotemporal regression problems using numerical weather predictions and physical models as additional inputs. The model enjoys good performances in terms of prediction. Our results show a decrease in nRMSE of 1.38\% compared to the state-of-the-art model for country-aggregated PV output prediction. 

This LRCN  model also demonstrates inference capability. NWP variables and physical models inputs have intricate and location-dependent correlations with PV power output. We showed that the CNN modules of the LRCN model can learn the geographic impact of different meteorological factors on the PV power prediction. Our occlusion sensitivity analysis validates our dense spatial approach and also allows us to better understand the dominant features as a function of location. It provides an interesting visualization tool for energy providers to estimate the effect of several weather variables of their PV power output. We also show that combining PVNet and a Sensitivity analysis can be used to learn an estimate of the surface density of PV power production for a given area.

Future work will involve refinements of the current model, introducing for example interpolation of NWP inputs in order to avoid temporal down-sampling and data loss. Then, we will build on the LRCN architecture by incorporating new spatial information, like satellite imagery or fish-eye data imagery.
A reliable PV forecast thus relies on models that fully leverage the available geographical data. With this work, we aim to open the doors to more spatiotemporal methods in PV output forecasting. 
The objective of such models is to enable a secure and economic integration of PV power into smart energy grids of countries over the world. This leads, in turn, to an increase in the proportion of renewable clean energies in the world's energy consumption. 


\bibliography{pvnet}
\bibliographystyle{icml2019}

\end{document}